\documentclass{article}

\usepackage{PRIMEarxiv}

\usepackage[utf8]{inputenc} 
\usepackage[T1]{fontenc}    
\usepackage{hyperref}       
\usepackage{url}            
\usepackage{booktabs}       
\usepackage{amsfonts}       
\usepackage{nicefrac}       
\usepackage{microtype}      
\usepackage{lipsum}
\usepackage{fancyhdr}       
\usepackage[pdftex]{graphicx}       
\usepackage[numbers]{natbib}
\usepackage{xcolor}

\usepackage{amsthm}
\usepackage{subcaption}
\usepackage{wrapfig}
\usepackage{multirow}
\usepackage{subfiles}
\usepackage{algorithm}
\usepackage{algorithmic}

\usepackage{amsmath,amsfonts,bm}

\def\eqref#1{equation~\ref{#1}}

\def\1{\bm{1}}

\DeclareMathAlphabet{\mathsfit}{\encodingdefault}{\sfdefault}{m}{sl}
\SetMathAlphabet{\mathsfit}{bold}{\encodingdefault}{\sfdefault}{bx}{n}

\newcommand{\R}{\mathbb{R}}

\DeclareMathOperator*{\argmax}{arg\,max}

\def\XX{\textsc{SPADE}}
\def\xx{Sample-efficient ProbAbilistic Detection using Extreme value theory}

\def\Y{\mathcal{Y}}

\def\D{\mathcal{D}}

\def\R{\mathbb{R}}

\def\CifarC{\text{CIFAR-100}}
\def\CifarD{\text{CIFAR-10}}
\def\ImageNet{\text{ImageNet-1K}}

\def\ResNet{\text{ResNet}}
\def\VIT{\text{ViT}}
\def\VGG{\text{VGG}}
\def\x{\mathbf{x}}

\def\z{\mathbf{z}}

\newcommand\norm[1]{\left\Vert#1\right\Vert}

\newtheorem{theorem}{Theorem}

\newtheorem{definition}{Definition}

\title{Provably Safeguarding a Classifier from OOD and Adversarial Samples: an Extreme Value Theory Approach\\~\\
\small{\textit{preprint}}}

\author{
  Nicolas Atienza\\
  Thales cortAIx-Labs \\
  Industrial AI Laboratory SINCLAIR\\
  Paris-Saclay University\\
  Palaiseau, France\\
  \And
  Christophe Labreuche\\
  Thales cortAIx-Labs \\
  Industrial AI Laboratory SINCLAIR\\
  Palaiseau, France\\
   \And
  Johanne Cohen, Michèle Sebag \\
  LISN CNRS-INRIA\\
  Paris-Saclay University\\
  Saclay, France\\
}

\defcitealias{hendrycks2018baseline}{1}
\defcitealias{liang2020enhancing}{2}
\defcitealias{lee2018simple}{3}
\defcitealias{pmlr-v162-sun22d}{4}
\defcitealias{ming2023exploithypersphericalembeddingsoutofdistribution}{5}

\def\Gc{\mbox{$\widehat{G}^{(c)}$}}

\def\Gs{\mbox{${G}^{(c)}$}}

\def\Gcc{\mbox{${G}^{(c,c')}$}}

\begin{document}
\maketitle


\begin{abstract}
This paper introduces a novel method, {\em Sample-efficient Probabilistic Detection using Extreme Value Theory} (\XX),  which transforms a classifier into an abstaining classifier, offering provable protection against out-of-distribution and adversarial samples. The approach is based on a Generalized Extreme Value (GEV) model of the training distribution in the classifier's latent space, enabling the formal characterization of OOD samples. Interestingly, under mild assumptions, the GEV model also allows for formally characterizing adversarial samples. The abstaining classifier, which rejects samples based on their assessment by the GEV model, provably avoids OOD and adversarial samples. The empirical validation of the approach, conducted on various neural architectures (ResNet, VGG, and Vision Transformer) and 
medium and large-sized datasets (CIFAR-10, CIFAR-100, and ImageNet), demonstrates its frugality, stability, and efficiency compared to the state of the art.
\end{abstract}


\section{Introduction}
A key challenge in deploying learned models in real-world settings is managing out-of-distribution (OOD) samples.  When a learned model encounters data that deviates from the training distribution, it can  lead to 
failures with significant consequences, particularly in high-stakes applications such as medical diagnosis, autonomous driving or risk analysis~\citep{salehi2022survey,yang2022openood}. 
The ultimate aim in machine learning  is to achieve OOD generalization, 
where the model 
encapsulates the core concept with sufficient accuracy to effectively handle atypical but real samples~\citep{ye2021theoretical}. A step towards OOD generalization is OOD detection, which equips the learned model with the ability to recognize atypical samples and refrain from making risky predictions. OOD detection is approached from several directions, including methods based on classification~\citep{hendrycks2018baseline,liang2020enhancing,hsu2021generalizedodindetectingoutofdistribution}, reconstruction metrics~\citep{jiang2023readaggregatingreconstructionerror,li2023rethinkingoutofdistributionooddetection}, density estimation~\citep{ren2019llr,liu2021energybasedoutofdistributiondetection,du2022voslearningdontknow} 
and distance-based estimation~\citep{papernot2018,pmlr-v162-sun22d,papernot2022,ming2023exploithypersphericalembeddingsoutofdistribution} (more in Section~\ref{sec:sota}). 

OOD detection is complicated by the fact that, to the best of our knowledge, no universally accepted definition exists
for what qualifies as an OOD sample. The boundaries between in-distribution 
and OOD data are inherently ambiguous and different domain experts may classify the same sample differently based on their understanding and experience~\citep{idrissi2022imagenetxunderstandingmodelmistakes}. Consequently,  the validation of OOD detection methods relies heavily on experimental studies using well-curated datasets,  such as near and far OOD datasets~\citep{yang2022openood}.

It is worth noting that human experts and models tend to make different decisions regarding both OOD and adversarial samples~\citep{goodfellow2015explaining}, albeit in distinct ways. Experts typically recognize an OOD sample as belonging to a given class, despite its atypicality, while the model assigns it to a random class. Conversely, experts perceive an adversarial example as typical of a specific class, yet the model confidently misclassifies it into a different class.

The approach presented in this paper, referred to as  {\em \xx} (\XX) and inspired by distance-based approaches~\citep{pmlr-v162-sun22d}, introduces an original model of the training distribution relative to a learned model (hereafter the teacher). Specifically, the distances between samples in the teacher's latent space are modelled using the Extreme Value Theory (EVT)~\citep{Fisher_Tippett_1928}. This model provides a sound and robust test for detecting and rejecting OOD samples. Most interestingly, under mild assumptions  this test also provably rejects adversarial examples with high probability, subject to a bound on their perturbation amplitude.  

The contributions of the proposed approach are fourfold: i) it introduces a formal definition of OOD samples relative to a teacher and its latent representation; ii) this definition leads to a statistically frugal OOD test based on EVT first principles; iii) this test operationally rejects OOD and adversarial samples with provable guarantees; iv) the effectiveness of the approach is experimentally and successfully demonstrated against strong baselines for learned models with different architectures~\citep{He2015deep,dosovitskiy2020vit,ming2023exploithypersphericalembeddingsoutofdistribution}.

The paper is organized as follows. Section~\ref{sec:formal_bg} outlines the formal background of distance-based OOD detection and introduces extreme value theory. Section~\ref{sec:overview} gives an overview of the \XX\ approach and its formal analysis. Sections~\ref{sec:exp_settings} and~\ref{sec:exp_results} respectively detail the experimental setup and the experiments conducted to validate \XX\ against state-of-the-art methods. Section~\ref{sec:sota} discusses the contributions within the context of related work, and the paper concludes with perspectives for future research.

\paragraph{Notations.} Let ${\cal D} = \{ (\x_1, y_1) \ldots (\x_n,y_n)\}$ denote the training set, iid drawn after the joint distribution $P_{X, Y}$, with ${\cal X}$ the instance space and ${\cal Y} = \{1, \ldots n_c\}$ the set of classes. The trained teacher $f$ is expressed as $f = g \circ h$, where $h$ is the embedding in the latent space ($h: {\cal X} \mapsto \R^d$) and $g$ is the mapping used to make decisions based on the latent representation. 

\section{Formal Background} \label{sec:formal_bg}
This section describes the main concepts defined by~\citep{ye2021theoretical} in the context of distance-based OOD generalization, and briefly introduces extreme value theory for completeness.

\subsection{Properties of Latent In-Distribution}
As mentioned above, the complexity of OOD characterization corresponds to the highly complex and diverse nature of real-world data~\citep{farquhar2022what}. In the literature \citep{cimpoi14describing,vanhorn2018inaturalist,haoqi2022vim,vaze2022openset,bitterwolf2023ninco} OOD characterization often involves subjective assessment (
"data that appear noticeably different from in-distribution to human observers"). A common definition states that the OOD is 
different from the known in-distribution, e.g., the training distribution,
although this definition does not capture the specifics of OOD: 
OOD is not merely any distribution that differs from ID.

On the other hand, as noted by~\cite{papernot2018,pmlr-v162-sun22d}, the representation in latent space of the in-distribution (hereafter ID) presents 
distinct characteristics, such as being formed of compact and well-separated clusters.~\cite{ye2021theoretical} formalize these properties in terms of variation and informativeness of the latent representation: 

\begin{definition}[\citep{ye2021theoretical}] \label{def:variation} The \emph{variation of embedding} $h$ across a finite distribution $\mathcal{D}$, noted $\mathcal{V}_\rho(h, \mathcal{D})$,
is defined as the maximum diameter over all classes $c$ of the ball containing distribution $h(\x)$ for $(\x,y) \in {\cal D}$ and $y=c$:
\begin{equation}
    \mathcal{V}(h, \mathcal{D}) = \max_{c \in Y} \sup_{\substack{(\x,c) \in {\cal D}\\ (\x',c) \in {\cal D}}} \|h(\x) - h(\x')\|
\end{equation}
where the distance $\|h(\x) - h(\x')\|$ is usually set to $L_2$ distance\footnote{The Kullback Leibler divergence is considered when embedding $h$ is a probabilistic one. See~\cite{ye2021theoretical} for more detail.}. Embedding $h$ is said $\eta$-invariant across $\cal D$ if 
$\mathcal{V}(h, \mathcal{D}) < \eta$.
\end{definition}
The variation of embedding $h$ thus measures the maximum thickness and width of the latent manifold containing the image of (samples in) a class. 

\begin{definition}[\citep{ye2021theoretical}] \label{def:informativeness} The \emph{informativeness of embedding} $h$ across a finite distribution $\mathcal{D}$,  noted $\mathcal{I}(h, \mathcal{D})$, is defined as the average over all pairs $(c,c')$ of distinct classes, of the minimum distance between $h(\x)$ and $h(\x')$ for $(\x,y)$ and $(\x',y')$ in $\cal D$, with  $y=c \neq y'=c'$:
\begin{equation}
    \mathcal{I}(h, \mathcal{D}) = \frac{1}{n_c(n_c-1)} \sum_{c\neq c' \in \Y} \min_{ \substack{(\x,y) \in {\cal D}, (\x',y') \in {\cal D}\\ y=c \neq y'=c'}} \|h(\x) -  h(\x')\|
\end{equation}
where the distance is usually set to $L_2$ distance (see footnote 1), and $n_c$ denotes the number of classes. The latent embedding $h$ is said \emph{$\delta$-informative} across $\cal D$ if 
$\mathcal{I}(h, \mathcal{D}) > \delta$.
\end{definition}

It is noted that informativeness and variation are closely related to the \emph{compactness} and \emph{dispersion} metrics introduced and optimized in CIDER to achieve 
OOD generalization~\citep{ming2023exploithypersphericalembeddingsoutofdistribution}. 

These definitions are defined in terms of supremum over classes, raising the challenge of their statistical stability and algorithmic exploitation (e.g., through setting thresholds). The approach presented here addresses this challenge by exploiting extreme value theory, as described below and referring the reader to~\cite{deHaan2006} for a comprehensive introduction.

\subsection{Extreme Value Theory} 
Dating back to~\cite{Fisher_Tippett_1928,Gnedenko43}, Extreme Value Theory (EVT) focuses on modeling and understanding the tail behavior of distributions. EVT is based on the premise that, under mild assumptions,
the distributions of extreme events converge to a common form, even if their original distributions differ.
 For instance, while the distributions of seismic intensities and the heights of rogue waves $-$ factors that respectively influence the design of buildings and oil rigs $-$ may differ, their extreme values (maxima) are governed by the same class of distributions. This limiting distribution is known as the \emph{Extreme Value Distribution} (EVD):

\begin{definition}[Extreme Value Distribution (EVD)~\citep{Fisher_Tippett_1928}] \label{def:evt} Let $Z$ be a random variable over the real-valued space $\R$. Let $Z^{(\ell)}$ denote the random variable defined as the maximum value over $\ell$ independent drawings of $Z$. When $\ell$ goes to infinity, the \emph{limiting distribution} of $Z^{(\ell)}$ is the cumulative distribution $ P(Z^{(\ell)} < z) \underset{\ell\to\infty}{\to} G_{\xi,\mu,\sigma}(z)$, expressed as one of the two parametric models:
\begin{equation}
G_{\xi,\mu,\sigma}(z) = \exp \left\{
\begin{array}{ll}
\left(  1+\xi \frac{z-\mu}{\sigma} \right)_+^{-1/\xi} & \mbox{ if } \xi \neq 0\smallskip\\
- \exp \left(\frac{\mu - z}{\sigma}\right)  & \mbox{otherwise}\end{array} \right \} 
\end{equation}
with $\mu\in\R$ a location parameter, $\sigma\in\R_+$ a dispersion parameter and $\xi\in\R$
a shape parameter referred to as \emph{extreme value index}. 
\end{definition}

Overall, the EVT framework provides a general parametric model for extreme events associated with a random variable, independent of the  distribution of $Z$ itself. The universality of these models reflects the fact that modeling the extreme events associated with a distribution relies only on the behavior of its tail. This tail can take one of three forms: (i) an exponential tail ($\xi=1$, corresponding to the Gumbel distribution); (ii) a heavy tail ($\xi>0$, corresponding to the Fréchet distribution); or (iii) a bounded tail ($\xi<0$, corresponding to the Weibull distribution).
Despite its applicability, EVT has, to the best of our knowledge, seen limited use in machine learning, with notable exceptions in the area of anomaly detection~\citep{Smith2012OnlineMA,siffer2017evtad,French2019QuantifyingTR}.

\section{\XX\ Overview} \label{sec:overview}
Aimed at OOD detection, \XX\ proposes a formal characterization of OOD concerning a trained teacher model and the associated latent representation on the one hand and the training distribution (hereafter in-distribution, ID) on the other hand. 
This characterization relies on generalized extreme value (GEV) models, which allow for detecting and rejecting out-of-ID samples. For simplicity and by abuse of language, out-of-ID samples are referred to as OOD in the following. Interestingly, under mild assumptions, the GEV models also allow for detecting adversarial samples. The abstaining classifier, equipped with the GEV-based detection tests, provides probabilistic guarantees of OOD and adversarial sample rejection, subject to a lower bound on the magnitude of the adversarial perturbation.

\subsection{EVT-based Characterization of OOD}
The distance-based OOD detection literature (see e.g., \citep{papernot2018,papernot2022,pmlr-v162-sun22d}) suggests that a sample is {\em likely} to be an OOD sample if it is {\em sufficiently} distant from the training samples of all (or most) classes in the latent space.

In \XX, this process is reformulated using generalized extreme value models, directly yielding the probability for a sample to be OOD. Let $(X, Y)$ denote the random variable following the joint distribution $P_{X,Y}$. 
For $Y=c$, let $Z_c$ be the random variable defined as the distance between $h(X)$
and its $k$-th nearest neighbor in latent distance, belonging to $\cal D$ with same class $c$. 
By definition, the limiting distribution of the maxima of $Z_c$ follows a  Generalized Extreme Value model noted $\Gs$, with $Pr(Z_c^{(l)} < v) \underset{\ell\to\infty}{{\to}} \Gs(v)$. 

For each sample $\x$ in the instance space and each class $c$, let $z_c$ be defined as $\|h(\x) - h(\x_{knn,c})\|$ with $\x_{knn,c}$ the $k$-th nearest neighbor of $\x$ in latent space, such that $(x_{knn,c},c)$ belongs to $\cal D$. 
As the true label of $\x$ is unknown at inference time, the proposed OOD test 
retains the lowest probability of $\x$ being OOD according to all $\Gs$:
\begin{definition}[OOD test]\label{def:ood}
Let $\x$ denote an instance in $\cal X$ with $z_c$ its Euclidean latent distance to its k-nearest neighbor of class $c$ in the training set $\cal D$. 
The probability of $\x$ being an OOD sample, noted $OOD(\x)$, is defined as:
\begin{equation}
 OOD(\x) = \min_{c \in {\cal Y}} \Gs(z_c)  
 \label{eq:OOD}
\end{equation}  
\end{definition}
In other words, $\x$ is considered to be OOD if it is extreme concerning all GEV models associated with the different classes.  

The decision to consider a separate GEV $\Gs$ for each class $c$ is intended to address situations where classes exhibit different levels of variation in the latent space. In such cases, considering a single GEV model for all classes could result in either erroneously rejecting samples from a class with high variation or incorrectly accepting OOD samples from a class with low variation.

Since the OOD test depends on the classifier's latent representation, one might wonder to what extent different tests based on different classifiers are consistent (as discussed further in Section~\ref{sec:exp_results}).

\subsection{Abstaining Classifier on OOD Samples}

A classifier abstaining on OOD samples is built as follows.

\begin{definition}[Abstaining classifier]
Given teacher $f$ and confidence $1-\tau$, with $0<\tau <1$, classifier $f_\tau$ abstains from making predictions on a sample $\x$ if $\x$ is considered to be extreme with probability at least $1-\tau$ w.r.t. its candidate class $c = f(\x)$. With 
same notations as above: 
\begin{equation}
f_{\tau}(\x) = \left \{ \begin{array}{ll} f(\x) & \mbox{if~} z_c \leq \Gs^{-1} (1-\tau) \\
\mbox{abstain} & \mbox{otherwise}\end{array} \right.   
\label{eq:abs}
\end{equation} 
where $z_c$ is the distance between $h(\x)$ and its nearest neighbor of class $c$ in $\cal D$.
\end{definition}

The abstention test embedded in $f_\tau$ is more precise than the OOD test (Def. \ref{def:ood}) because it incorporates the additional information of $f(\x)$. 
Note, however, that both tests coincide under the common assumption that an OOD sample is closer to the samples of its own class, all else being equal.

\subsection{Abstaining classifier with provable guarantees w.r.t. adversarial examples}
Let us consider the adversarial example $\x$ built by perturbing a training sample noted $\x^*$ of class $c$, and let $c' = f(\x) \neq c$ be its (wrong) class according to $f$. 
Under mild assumptions, 
it is shown that the abstaining classifier $f_\tau$ abstains on adversarial sample $\x$ with probability $1 - \tau$, subject to a lower bound on its perturbation amplitude.

Let $\Gcc$ denote the generalized extreme value model associated with the {\em minimum} latent distance among pairs of examples $(\x,\x')$  respectively belonging to class $c$ and $c'$: 
\begin{equation*}
  \Gcc(v) = Pr \Bigl(\|h(X) - h(X')\| > v~\big|~ (X,Y) \sim P_{X,Y}, ~(X',Y') \sim P_{X,Y}, ~Y=c,~ Y'=c'\Bigr)   
\end{equation*} 

\begin{theorem}
Let us assume that the latent embedding $h$ is  $K$-Lipschitz.
Let $\x$ be an adversarial sample built by perturbation of a training sample $\x^*$ of class $c$, with perturbation amplitude $\varepsilon$  ($\|\x - \x^*\| < \varepsilon$), and let $f(\x) = c' \neq c$. Let $\x'^*$ of class $c'$ denote the $k$-th nearest training sample in $\cal D$ of $\x$. 
Then, with probability at least $1 - \tau$ either $f_\epsilon$ abstains on $\x$, or perturbation $\epsilon$ is greater than the following lower bound:
\begin{equation}
\varepsilon \ge \frac{1}{K} \left( \Gcc^{-1}(1-\tau) - \Gs^{-1}(1-\tau) \right)
\label{EqTh1}
\end{equation}
\end{theorem}
\begin{proof}
If $\|h(\x) - h(\x'^*)\| > (\Gs)^{-1}(1- \tau)$, then $f_\tau$ abstains on $\x$ (Def. 5).\\
Otherwise, 
\begin{align}
 \|h(\x^*) - h(\x'^*)\| & \leq \|h(\x^*) - h(\x)\| + \|h(\x) - h(\x'^*)\|  \nonumber \\
 & \leq K \varepsilon + (\Gs)^{-1}(1-\tau)
\label{EqTh2}
 \end{align}
Moreover, with probability $1-\tau$, 
 \begin{align} (\Gcc)^{-1}(1 - \tau) \leq  \|h(\x^*) - h(\x'^*)\|
 \label{EqTh3}
 \end{align}
 Putting together Eqs.~\ref{EqTh2} and~\ref{EqTh3} concludes the proof.
\end{proof}

\subsection{Estimating the GEV Models}
The estimation of the GEV models involved in \XX\ is detailed in the case of $\Gs$, defining its approximation $\Gc$. The same process is used to learn an approximation of the $\Gcc$ models, with the only difference being that the considered extreme values are minima instead of maxima.

After \cite{siffer2017evtad}, a straightforward approximation of an EVD proceeds by sampling the extreme events along the block maxima method. This fitting process is, however, found to be sensitive to the number of blocks and the block size. 
The proposed approach thus is the \emph{Peak Over Threshold} (POT) method, which relies on the Pickands-Balkema-de-Haan theorem, often referred to as the \emph{second theorem of EVT} \citep{balkema74-pot,pickands75pot}. Formally, the POT considers the occurrences of events bypassing a threshold $t$, noting that the distribution of these occurrences follows a Generalized Pareto Distribution (GPD):
\begin{equation}
    F_{\xi,\mu,\sigma}(z) = Pr (Z-t>z | Z>t ) = \left\{ \begin{array}{ll}
          1- \exp\left(-\frac{z-\mu}{\sigma}\right) \hspace{.3 cm} & \mbox{if } \xi=0  \\
          1 - \left( 1 + \frac{\xi (z-\mu)}{\sigma}\right)^{-1/\xi} \hspace{.3cm} & \mbox{otherwise}
         \end{array} \right.
\end{equation} 
Informally, for each class $c$, POT proceeds by fitting a GPD model to samples $z_c$ that are over a threshold $t_c$. Formally, for each sample 
$(\x_i,c)$ in  $\cal D$, let $z_i \in \R$ be the distance of $\x_i$ to its $k$-th nearest neighbor belonging to class $c$, in latent space. Let ${\cal D}_c$ be the set including all such $z_i$ for $z_i > t_c$. 

The parameters $(\mu_c,\sigma_c,\xi_c)$ of model $\Gc$ are learned by maximum likelihood estimation (MLE) on ${\cal D}_c$:
\begin{equation}
    (\mu_c,\sigma_c,\xi_c) =  \argmax\limits_{\mu,\sigma,\xi} \sum_{z \in {\cal D}_{c}} {\cal L}_{\mu,\sigma,\xi}(z) 
\end{equation}
with $\cal L$ the log-probability density function of the GPD. MLE is preferred over alternative methods, e.g. the method of moments, due to its higher robustness and efficiency \citep{siffer2017evtad}.

Eventually, model $\Gc$ approximately characterizes whether a given sample is OOD with respect to class $c$ with confidence $1 - \tau$. The OOD test (Def. \ref{def:ood}) is accordingly approximated as: 
\begin{equation}
 \widehat{OOD}(\x) =  \min_{c \in {\cal Y}} \Gc(z_c)  
 \label{eq:OOD-approx}
\end{equation}  

\subsection{Discussion} 
\XX\ retains the main benefits of distance-based OOD detection approaches, being agnostic to the structure of the OOD distribution and easy to implement. Furthermore, being grounded in the EVT first principles, it enables estimating the probability for a sample to be OOD. Lastly, the abstaining classifier $f_\tau$ can also reject adversarial samples, subject to a lower bound on the magnitude of adversarial perturbations. 

A potential weakness is the complexity of approximating GEV models, which is quadratic with respect to the number $n$ of samples. This raises the question of whether stable and accurate GEV approximations can be achieved when aggressively subsampling the training set. A second issue pertains to the effectiveness of the lower bound used in rejecting adversarial samples (Eq. \ref{EqTh1}); specifically the question is whether $\Gcc^{-1}(1 - \tau) - \Gs^{-1}(1 - \tau)$ is strictly positive for non-trivial confidence levels $1 - \tau$.\footnote{ Note that the requirement for $(\Gcc)^{-1}(1 - \tau)$ to be sufficiently large is reminiscent of the clustering assumption that underpins semi-supervised learning \citep{cluster-assumption}.}
The \XX\ method can be summarized in the algorithm below: 

\begin{algorithm}[H]
\caption{\XX. Learning EVT models}
\label{alg:spade}
\begin{algorithmic}[1]
\STATE \textbf{Input:} training set $\D$, integer $k$, threshold $t > 0$
\STATE \textbf{Output:} GEV distributions $(\Gc)$ of each class $c$ 
\STATE \FOR{$c \in \Y$}
\STATE $\mathcal{T}^{(c)} = \{\}$
\FOR{$(\x,c) \in \D$}
\STATE Compute normalized activation: $\z = h(\x)/\norm{h(\x)}$ 
\STATE Compute distance to $k$-th nearest neighbor in class $c$: $v = \min\limits^{(k)}_{(\x',c)\in\D}\left(\norm{\z-\z'}\right)$
\STATE \textbf{if} $v>t$: $\mathcal{T}^{(c)} = \mathcal{T}^{(c)} \cup \{v\}$ \textbf{end if}
\ENDFOR
\STATE Fit an Extreme Value Distribution (\Gc) from extreme samples of $\mathcal{T}^{(c)}$
\ENDFOR
\RETURN $(\Gc)_{c\in\Y}$ 
\end{algorithmic}
\end{algorithm}

\section{Experimental Setting} \label{sec:exp_settings}
This section outlines the experimental setup used to evaluate SPADE in comparison to the state of the art. All experiments were conducted on Tesla A100 80GB GPUs. Further details are provided in the supplementary material (SM).

\paragraph{Goals.} The experiments aim to empirically address four questions: the performance of the OOD detection tests based on the GEV models, particularly in comparison with distance-based approaches (Q1); their sensitivity with respect to the considered teacher (Q2); similarly, the performance of the adversarial sample detection test based on the GEV models, compared with state-of-the-art methods (Q3);  the computational complexity and stability of the approximate GEV models embedded in \XX\ (Q4). 

The performance of \XX\ is compared to five established baselines: the seminal MSP \citep{hendrycks2018baseline}, ODIN \citep{liang2020enhancing}, MDS \citep{lee2018simple}, KNN \citep{pmlr-v162-sun22d}, and its extension CIDER \citep{ming2023exploithypersphericalembeddingsoutofdistribution} (further discussed in Section~\ref{sec:sota}). 

\paragraph{Metrics.} The comparative evaluation is conducted using the OpenOOD framework \citep{zhang2023openood}, with performance assessed by standard indicators: The \emph{Area Under the ROC Curve (AUC)} measures the average rate of correct OOD/adversarial sample detection across all confidence levels $2 - \tau$, corresponding to the true positive rate, as described in \citep{zhang2023openood}. The \emph{FPR95 indicator} represents the fraction of true samples misclassified as OOD (respectively, adversarial) when the detection threshold ensures 95\% of OOD (resp., adversarial) samples are correctly rejected.

\paragraph{Benchmarks.} Three medium- and large-sized datasets are considered: \CifarD, \CifarC\ \citep{Krizhevsky2009LearningML} and \ImageNet\ (using the ILSVRC2012 version). 
Three types of neural architectures are used to assess the generality of the \XX\ approach: \ResNet\ \citep{He2015deep}, \VIT\ \citep{dosovitskiy2020vit}, and \VGG\ \citep{simonyan2014very}. The distance in latent space between a sample and its $k$-th nearest neighbor is calculated using the normalized $L_2$ distance, following \cite{pmlr-v162-sun22d}. OOD samples are sourced from near-OOD and far-OOD datasets, following \citep{cimpoi14describing,vaze2022openset,vanhorn2018inaturalist,haoqi2022vim,bitterwolf2023ninco}, as detailed in the supplementary material (SM). Adversarial samples are generated by perturbing training samples using FGSM \citep{goodfellow2015explaining} attacks, with a perturbation amplitude $\varepsilon$ ranging from 0.001 to 0.004.

\begin{table}
    \centering
        \caption{OOD detection on \ImageNet: performance of \XX\ (with ResNet teacher) compared to that of baselines MSP, ODIN, MDS and KNN  in terms of AUC (the greater the better) and  FPR95 (the lower the better; best performances in bold). The rank, averaged over far and near OOD datasets, is computed after the half sum of AUC and FPR95.
        }
    \label{tab:OOD_compar}
    \resizebox{\linewidth}{!}{\begin{tabular}{ccccc|cccccc|c}
    \toprule
        &\multicolumn{4}{c}{\bf Near OOD} & \multicolumn{6}{|c|}{\bf Far OOD} & \multirow{3}{*}{{\bf Rank}} \\
       & \multicolumn{2}{c}{\bf SSB Hard} & \multicolumn{2}{c}{\bf NINCO} & \multicolumn{2}{|c}{\bf iNaturalist} & \multicolumn{2}{c}{\bf Textures} & \multicolumn{2}{c|}{\bf OpenImages-O} &   \\
       & \textbf{AUC} $\uparrow$ & \textbf{FPR95} $\downarrow$ &\textbf{AUC} $\uparrow$ & \textbf{FPR95} $\downarrow$ &\textbf{AUC} $\uparrow$ & \textbf{FPR95} $\downarrow$ &\textbf{AUC} $\uparrow$ & \textbf{FPR95} $\downarrow$ &\textbf{AUC} $\uparrow$ & \textbf{FPR95} $\downarrow$ &  \\
        \midrule
        MSP \citepalias{hendrycks2018baseline} &\bf 72.53&7\bf 4.43&\bf 80.66&\bf 57.72&87.78&44.08&82.81&59.16&85.21&49.62& \bf 3 \\
        ODIN \citepalias{liang2020enhancing} & 72.51 &77.36&77.55&70.83&\bf 89.51&\bf 41.46&87.02&56.58&86.33&54.10 & \bf 4\\
        MDS \citepalias{lee2018simple} &52.15&90.46&68.49&71.66&76.49&56.07&94.11&27.07&77.68&59.66& \bf 5\\
        KNN \citepalias{pmlr-v162-sun22d} &62.80&84.08&79.30&58.92&84.62&42.39&\bf 96.06&\bf 23.39&\bf 86.38&\bf 44.24& \bf 1\\ \midrule
        {\bf \XX} & 61.91 & 85.27 & 77.99 & 61.04 & 85.26 & 44.84 & 95.86 & 24.63  & 85.79 &  46.33 & \bf 2\\ 
        \bottomrule
    \end{tabular}}
\end{table}

\begin{table}
    \centering
     \caption{OOD detection on \CifarD: performance of \XX\ with teachers ResNet, VGG and ViT-B16, compared to that of baseline CIDER.}
    \label{tab:OODtest}
    \resizebox{\linewidth}{!}{\begin{tabular}{ccccccccccc}
    \toprule
       & \multicolumn{2}{c}{\bf TIN} & \multicolumn{2}{c}{\bf MNIST} & \multicolumn{2}{c}{\bf SVHN} & \multicolumn{2}{c}{\bf Textures} & \multicolumn{2}{c}{\bf Places-365}   \\
       & \textbf{AUC} $\uparrow$ & \textbf{FPR95} $\downarrow$ &\textbf{AUC} $\uparrow$ & \textbf{FPR95} $\downarrow$ &\textbf{AUC} $\uparrow$ & \textbf{FPR95} $\downarrow$ &\textbf{AUC} $\uparrow$ & \textbf{FPR95} $\downarrow$ &\textbf{AUC} $\uparrow$ & \textbf{FPR95} $\downarrow$  \\
        \midrule
        CIDER ($d$=512) &71.56 & 70.34 & 68.84 & 71.86& 57.51 & 78.43 & 71.06 & 70.70 & 71.73 & 69.97 \\
        \midrule
        ResNet-18 ($d$=512) &90.41&32.14&93.646&21.91&92.17&22.94&91.97&25.68& 91.03&30.41 \\
        VGG-16 ($d$=512) & 76.04 & 66.84 & 89.37 & 35.12 & 81.53 & 44.06 & 80.47 & 50.73 & 72.14 & 77.27\\
        ViT-B16 ($d$=384) & 96.27 & 20.18  &  94.52 & 12.23 & 81.78 & 34.83 & 99.97 & 00.06 & 99.67 & 00.51  \\
        \bottomrule
    \end{tabular}}
    \end{table}

\section{Experimental Results} \label{sec:exp_results}
\paragraph{OOD Detection (Q1).} The performance of \XX\ is illustrated in Table~\ref{tab:OOD_compar}, focusing on the representative case of \ImageNet\ and considering a ResNet teacher. 
For the considered near-OOD datasets, the best method is MSP, whereas for far-OOD datasets, the best method is KNN. In all cases but one, \XX-ResNet is slightly outperformed by KNN. In terms of rank (determined by the average of AUC and FPR95), \XX\ ranks second best on both near- and far-OOD datasets.

\paragraph{Sensitivity of OOD Detection (Q2).} The impact of the considered teacher on the performance of the \XX\ OOD test is illustrated in Table~\ref{tab:OODtest}, comparing \XX\ built on teachers ResNet, VGG, ViT-B16 with the CIDER baseline on \CifarD. These results show that the detection accuracy indeed depends on the teacher and its latent space, with \XX-ViT-B16 dominating ResNet (except on SVHN) and ResNet strongly dominating VGG. Still, the performance does not merely depend on the size of the latent space. The discrepancy between the AUC and FPR95 indicators suggests that the optimal detection threshold varies depending on the dataset. Notably, all \XX\ OOD detection tests outperform CIDER, which is based on KNN and specifically targets OOD detection. We shall come back to this remark in Section~\ref{sec:sota}.
\vspace*{-1.2\baselineskip}
\begin{wrapfigure}[25]{r}{0.36\linewidth}
    \centering
    \subcaptionbox{Extreme Value Index $\xi$}{\includegraphics[width=\linewidth]{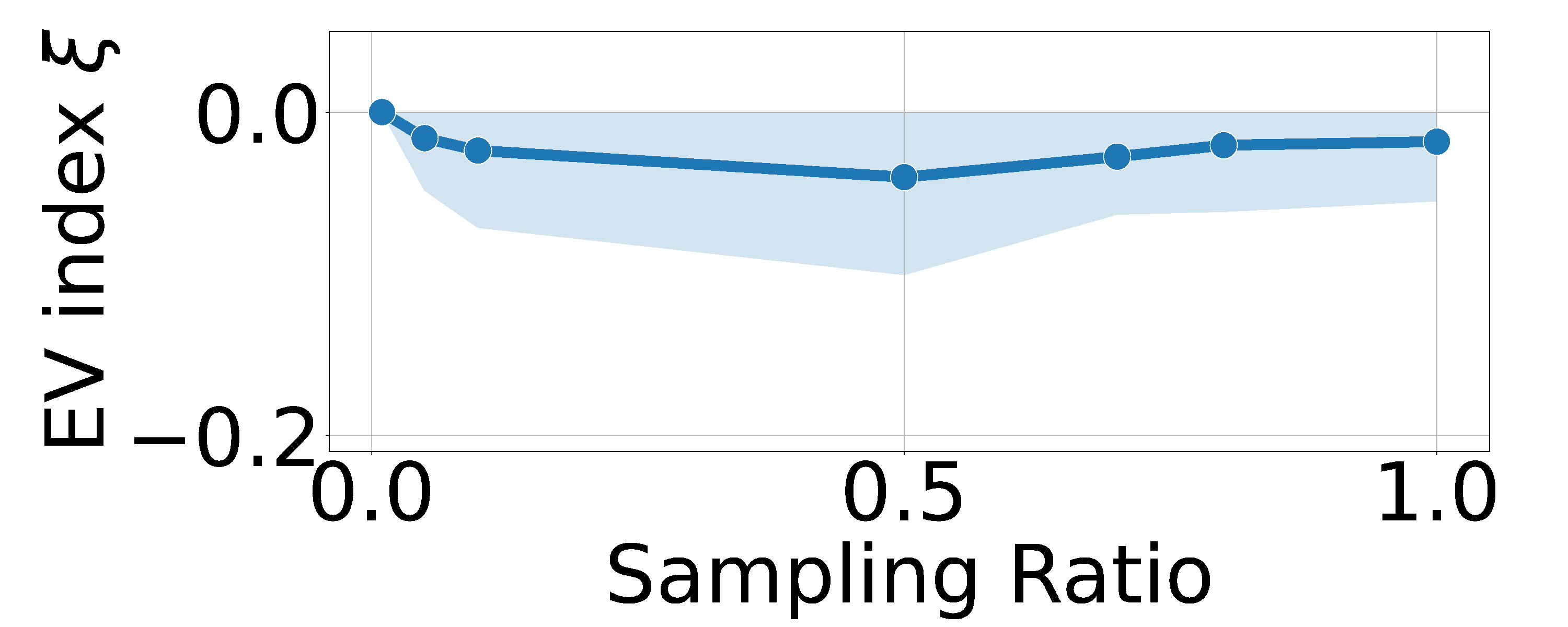}}
    \subcaptionbox{Location parameter $\mu$}{\includegraphics[width=\linewidth]{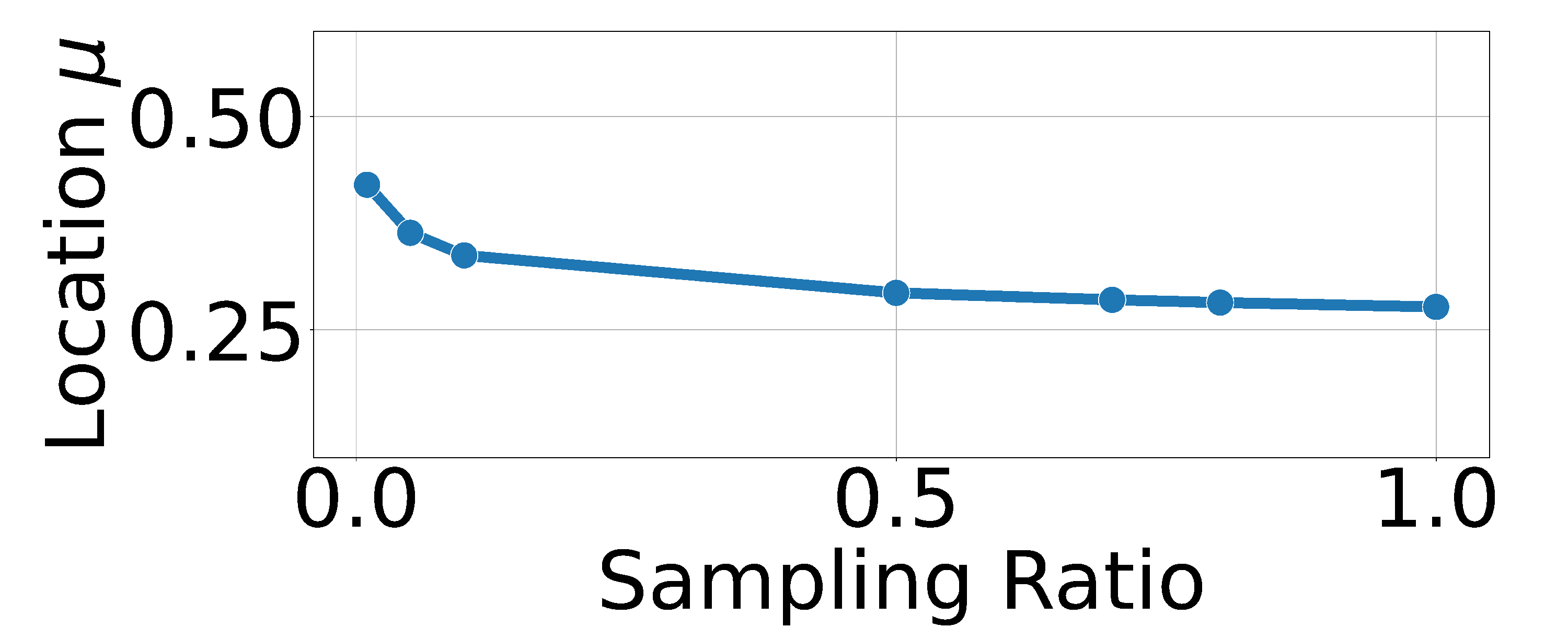}}
    \subcaptionbox{Digression parameter $\sigma$}{\includegraphics[width=\linewidth]{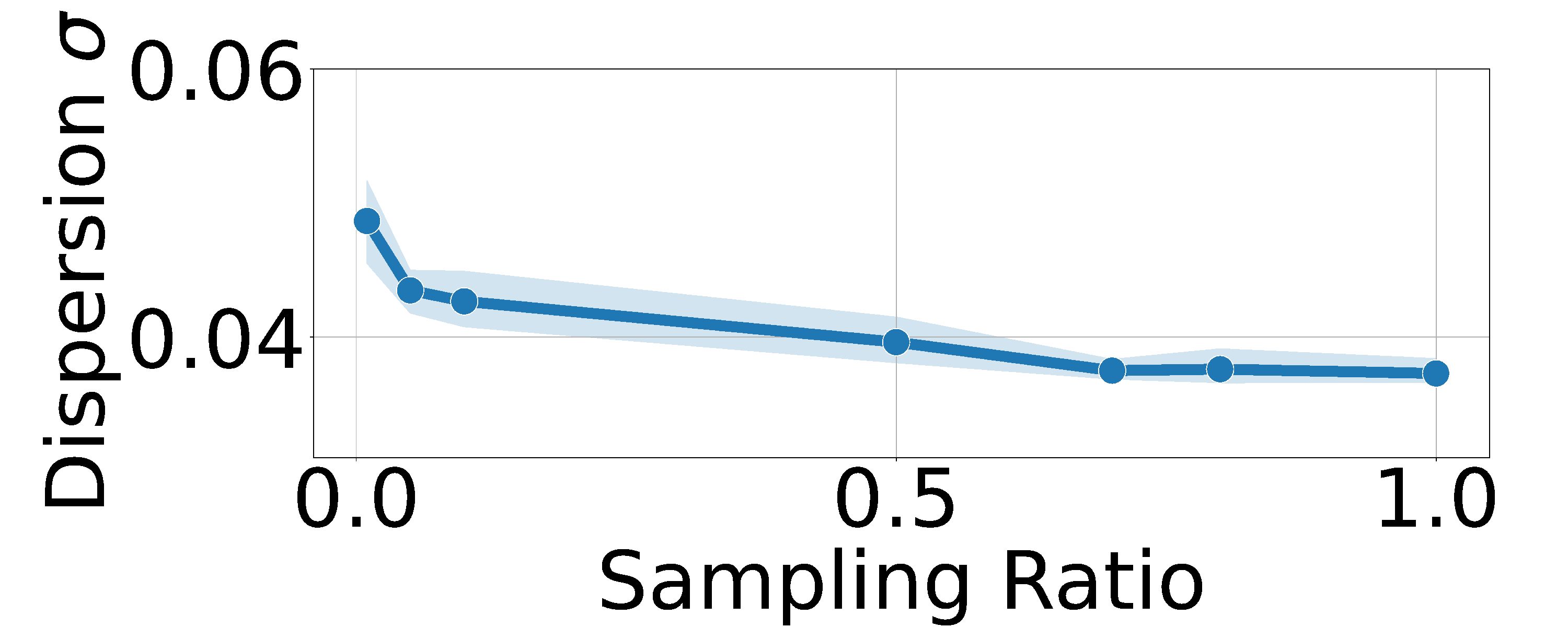}}
    \caption{Stability of EVT parameter estimation wrt sampling ratio and estimation variance for one class of \CifarC\ on ResNet-18.}
    \label{fig:evt-param}
\end{wrapfigure}
\mbox{}

\paragraph{Detection of Adversarial Samples (Q3).} Table \ref{tab:ood_adv} reports  the performance of \XX\  (with a ResNet teacher) on the representative cases of \CifarD\ and \CifarC, compared with MSP, MDS, KNN and CIDER. For all perturbation amplitudes, \XX\  ranks first w.r.t. AUC (and second w.r.t. FPR95). In terms of FPR95, KNN ranks first on \CifarD, while MSP ranks first on \CifarC. 
Overall, \XX\ behaves on par with,  or better than, OOD detection methods w.r.t. the detection of adversarial examples. 
The slight AUC improvement suggests that \XX\ may capture more subtle differences between in-distribution and adversarial samples. Conversely, the high FPR95 values suggest that \XX\ tends to be overly cautious, rejecting true samples at the level of confidence where 95\% adversarial samples are rejected.
\begin{table}
    \centering
        \caption{Adversarial samples detection on \CifarD\ and \CifarC, with perturbation amplitude from .001 to .004: comparison of \XX\ (ResNet teacher) with baselines MSP, ODIN, MDS and KNN  in terms of AUC (the greater the better) and  FPR95 (the lower the better; best in bold).}
        \label{tab:ood_adv}
    \resizebox{\linewidth}{!}{\begin{tabular}{cccccccccc|cc}
    \toprule
    && \multicolumn{2}{c}{$\varepsilon=0.001$} & \multicolumn{2}{c}{$\varepsilon = 0.002$} & \multicolumn{2}{c}{$\varepsilon = 0.003$} & \multicolumn{2}{c}{$\varepsilon = 0.004$} & \multicolumn{2}{|c}{Average} \\
    && \textbf{AUC} $\uparrow$ & \textbf{FPR95} $\downarrow$ &\textbf{AUC} $\uparrow$ & \textbf{FPR95} $\downarrow$ &\textbf{AUC} $\uparrow$ & \textbf{FPR95} $\downarrow$ &\textbf{AUC} $\uparrow$ & \textbf{FPR95} $\downarrow$ &\textbf{AUC} $\uparrow$ & \textbf{FPR95} $\downarrow$ \\
    \midrule 
\parbox[t]{2mm}{\multirow{5}{*}{\rotatebox[origin=c]{90}{\CifarD}}}&MSP \citepalias{hendrycks2018baseline}&81.68&79.08&81.90&78.70&82.06& 79.14&82.20&78.10&81.96&78.76 \\
&MDS \citepalias{lee2018simple}&81.46&69.34&81.51&68.76&81.57&69.04&81.61&69.42&81.54&69.14 \\
&KNN \citepalias{pmlr-v162-sun22d} &85.65&\bf 54.46&85.75&\bf 54.19&85.85&\bf 53.91&85.92&54.29&85.79&\bf 54.21 \\
& CIDER \citepalias{ming2023exploithypersphericalembeddingsoutofdistribution} &85.46&55.68&85.50&54.79&85.53&54.93&85.60&54.57&85.52&54.99 \\
 \cmidrule{2-12}
&\bf \XX &\bf 85.96&55.02&\bf 86.06&54.51&\bf 86.15&54.40&\bf 86.24&\bf 53.78&\bf 86.10&54.43 \\
    \midrule
     \parbox[t]{2mm}{\multirow{5}{*}{\rotatebox[origin=c]{90}{\CifarC}}}&MSP \citepalias{hendrycks2018baseline} &83.24&\bf 51.84&83.39&\bf 50.57&83.54&\bf 49.94&83.68&\bf 49.64& 83.46 &\bf 50.50\\
    &MDS \citepalias{lee2018simple} &60.34&82.93&60.29&83.01&60.25&82.60&60.19&83.01&60.27& 82.89\\
    &KNN \citepalias{pmlr-v162-sun22d} &83.54&56.83&83.67&56.39&83.79&55.64&83.89&55.13&83.72&56.00 \\
& CIDER \citepalias{ming2023exploithypersphericalembeddingsoutofdistribution} &82.75&63.17&82.85&63.57&82.95&63.17&83.06&62.34&82.90&63.06 \\
 \cmidrule{2-12}
    &\bf \XX &\bf 84.33&53.13&\bf 84.45&52.84&\bf 84.56&52.44&\bf 84.66&52.44&\bf 84.50& 52.72\\ 
\bottomrule
    \end{tabular}}
    \end{table}

\paragraph{Computational Frugality:  Stability of GEV Models and \XX\ Performances (Q4).} The stability of the GEV models with respect to the fraction of training samples used to estimate the \Gc\ hyperparameters is illustrated in Fig. \ref{fig:evt-param}, using the ResNet teacher's latent space on \CifarC. The figure shows: i) a low sensitivity of the tail index (Fig. \ref{fig:evt-param}.(a)); ii) a decrease in $\mu$ with the sampling rate (Fig. \ref{fig:evt-param}.(b)); iii) a low sensitivity of the dispersion parameter $\sigma$ (Fig. \ref{fig:evt-param}.(c)). Complementary results are provided in the SM, confirming that the lower bound on the adversarial perturbation amplitude is effectively positive. 

\begin{figure}
    \centering
    \subcaptionbox{near OOD}{\includegraphics[width=0.45\linewidth]{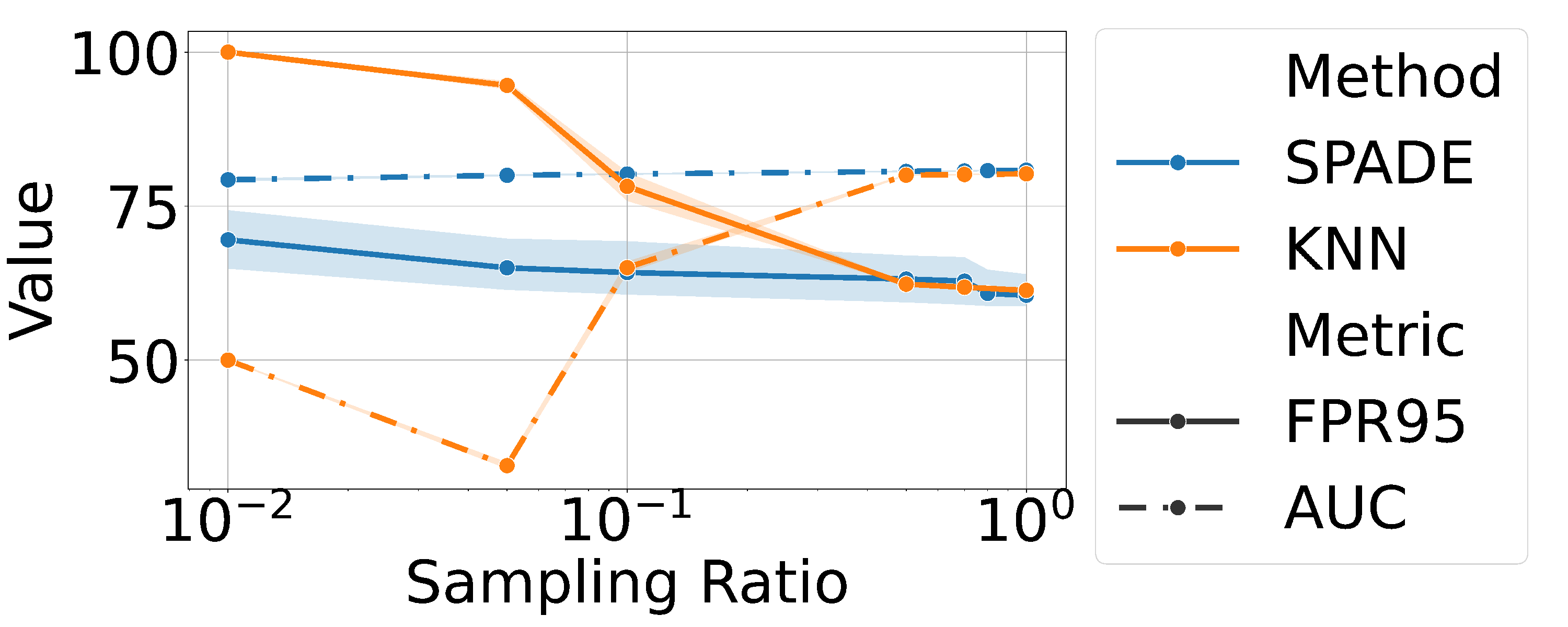}}
    \subcaptionbox{far OOD}{\includegraphics[width=0.45\linewidth]{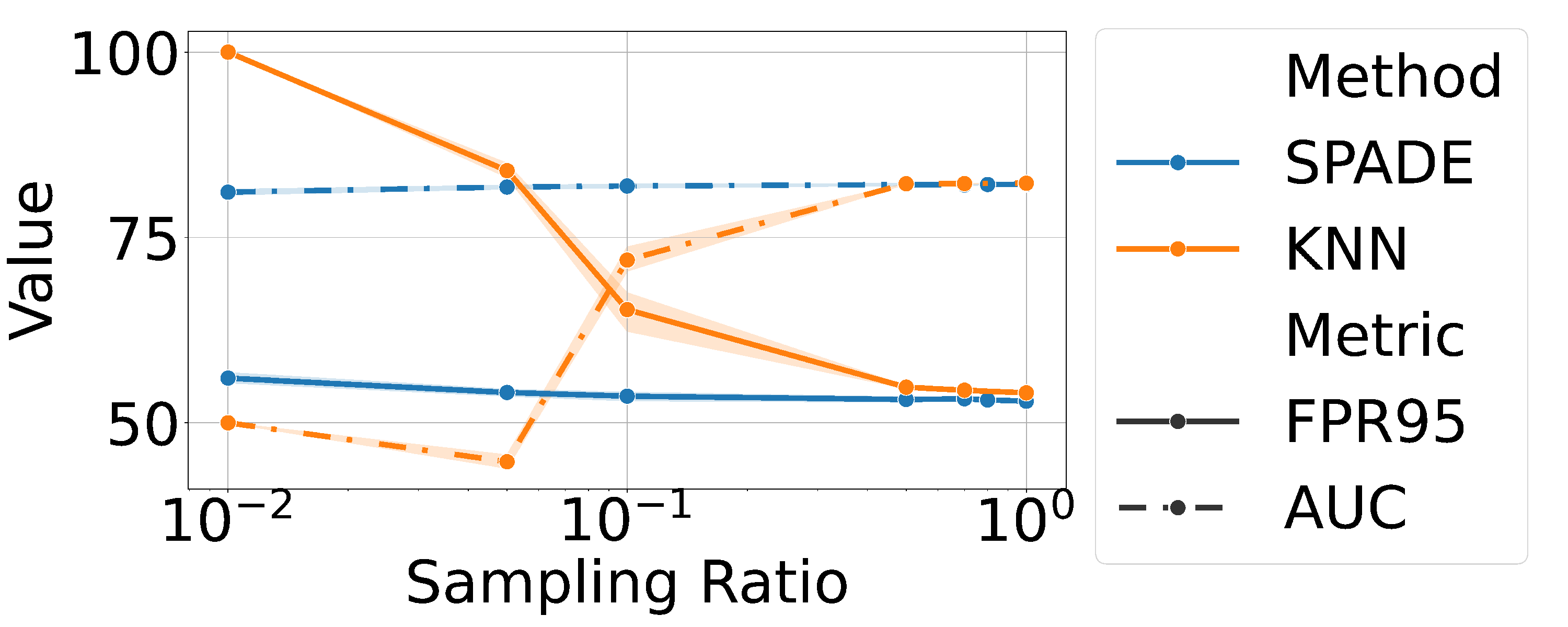}}
    \caption{Sensitivity analysis of the OOD detection on \CifarC\ w.r.t. the subsampling rate of the training set: AUC (dashed line) and FPR95 (solid line) performances
    for \XX\ (in blue) and KNN \citep{pmlr-v162-sun22d}  (in orange; better seen in color). }
    \label{fig:sample-complexity}
\end{figure}

As expected, the stability of the $\Gc$ models results in stable OOD detection performances (AUC and FPR95) when subsampling the training set. Quite the contrary, the OOD detection performances of KNN are significantly deteriorated when aggressively subsampling the training set, particularly so on near-OOD (Fig. \ref{fig:sample-complexity}).

\section{Position w.r.t. related work} \label{sec:sota}
The robustness lack of machine learning models with respect to adversarial and OOD examples is widely recognized as a major obstacle for ML applications in safety-critical domains \citep{salehi2022survey,yang2024survey}.

With respect to OOD examples, their detection takes inspiration from several areas of ML, ranging from learning with rejection~\citep{bartlett2008lwr}, to anomaly detection \citep{bulusu2020adsurvey}, novelty detection \citep{markou2003nd1,markou2003nd2}, and open set recognition \citep{Boult2019osr}. To the best of our knowledge, the OOD detection problem was first formalized by \citep{nguyen2015deep}; the early MSP approach, observing the margin between the logits of the trained teacher and exploiting the fact that it behaves differently for in-distribution and OOD samples \citep{hendrycks2018baseline}, still is among the most effective ones (Table \ref{tab:OOD_compar}). Along the same line, ODIN exploits the gradient information to separate in- and out-of-distribution samples  \citep{liang2020enhancing}. \\ 
Some approaches address OOD detection as yet another supervised learning issue, treating OOD samples as belonging to an additional class and utilizing them to (re-)train the model \citep{du2022unknownawareobjectdetectionlearning,zhang2022mixtureoutlierexposureoutofdistribution} (see also \citep{hsu2021generalizedodindetectingoutofdistribution}). 

Quite a few other methods are based on the so-called manifold assumption; the challenge, then, is to identify the representation that best characterizes the manifold on which real samples lie. One option is to consider the latent representation of an auto-encoder (AE) trained solely on real samples.
As empirically shown by e.g. \cite{ren2019llr,liu2021energybasedoutofdistributiondetection,du2022unknownawareobjectdetectionlearning}, the reconstruction error of OOD samples through the AE is much higher than for real samples; an effective OOD test can thus be based on this error. On the positive side, this reconstruction error defines a general criterion and does not depend on the teacher; on the negative side, it does not leverage class information.
Another option is based on modeling the behavior of true samples in the penultimate layer of the neural net, using probabilistic models \citep{du2022voslearningdontknow,liu2021energybasedoutofdistributiondetection,ren2019llr}.

Finally, the option most closely related to \XX\ is to consider the latent representation of the teacher itself, as is done in distance-based OOD detection approaches \citep{papernot2018,lee2018simple}, particularly in KNN \citep{pmlr-v162-sun22d}. The difference is that KNN exploits the distance  $z$ of a sample to its $k$-th nearest neighbor in the training set, while \XX\ exploits $\Gc(z_c)$. A tentative interpretation for the better performance of KNN (Table \ref{tab:OOD_compar}) is the superior bias-variance trade-off in the empirical test based on $z$ compared to the test based on $\Gc(z_c)$. While learning a parametric GEV model achieves some regularization, this model is only trained from examples within the class $c$, meaning it operates with one or two orders of magnitude less data.

The problem of dealing with adversarial examples differs from that of detecting OOD examples, as adversarial examples are deliberately crafted to deceive the teacher \citep{szegedy2014intriguing, goodfellow2015explaining, madry2019deep, croce2020aa}. Knowing their structure allows for the design of specific defense strategies, such as adversarial training \citep{goodfellow2015explaining, madry2019deep}, which incorporates adversarial examples into the training process \citep{NEURIPS2018_8f121ce0,pmlr-v97-zhang19p,Wang2020ImprovingAR}. Other notable defense strategies include adversarial architectures \citep{Hosseini2020DSRNADS, 10204909}, adversarial regularization \citep{NEURIPS2019_c24cd76e, LIU2023103877}, and data augmentation methods \citep{NEURIPS2019_32e0bd14, pmlr-v202-wang23ad}.

In contrast, \XX\ neither requires additional information nor retrains the classifier to defend against adversarial examples. It employs the same agnostic strategy for both adversarial and OOD threats: it characterizes the true samples and abstains from making a decision if the example in question appears extreme compared to the true (ID) samples, based on the chosen confidence level.

\section{Conclusion and Perspectives} \label{sec:conclusion}
The main contribution of the paper, \XX\ ({\em \xx}), is a formal test designed to detect examples appearing to be extreme w.r.t. training samples, enabling the classifier to abstain from making decisions on such extreme examples. This test's ability to accurately detect OOD and adversarial samples has been empirically demonstrated, with similar performances as the prominent state of the art approaches. Furthermore, the computational complexity and the stability of the proposed detection test have been empirically established. 

The proposed test, similar to distance-based OOD detection approaches, exploits the latent distance between the given example and its nearest neighbor in the training set. Its originality lies in leveraging  Extreme Value Theory \citep{Fisher_Tippett_1928} to provide a formal characterization of the training samples. This characterization offers two key benefits: first, it yields provable guarantees for detecting adversarial examples, subject to  the adversarial perturbation amplitude to be lower bounded; second, it provides some new hints  into the key aspects governing the teacher robustness.
\footnote{For instance, CIDER \cite{ming2023exploithypersphericalembeddingsoutofdistribution} involves the optimization of the variation and informativeness of the teacher latent space. The \XX\ analysis suggests that besides these two factors, the regularity of the teacher (its Lipschitz constant) also matters. A possible interpretation for why \XX\ outperforms CIDER is that the optimization of the variation and informativeness might adversely affect this Lipschitz constant.}

This approach opens several perspectives, related with making classifiers more robust and better understanding the key robustness factors. 
A short term perspective is to extend the generalized extreme value test to some of the empirical criteria used in the OOD detection literature; one such criterion is the score margin involved in MSP \citep{hendrycks2018baseline}.
Another perspective is to enhance the classifier training loss to favor the robustness of the latent space, e.g. to consider the optimization of the Lipschitz constant of the classifier embedding besides its variation and informativeness as done in CIDER \citep{ming2023exploithypersphericalembeddingsoutofdistribution}. 

Our long-term goal is to investigate whether {\em safe} example behaviors can be identified in the latent space and whether these behaviors can be certified, as a step toward the certification of neural networks.

\bibliographystyle{unsrt}  
\bibliography{main}

\end{document}